# tmn at #SMM4H 2023: Comparing Text Preprocessing Techniques for Detecting Tweets Self-reporting a COVID-19 Diagnosis

Anna Glazkova
University of Tyumen, Tyumen, Russia

**Abstract**
*The paper describes a system developed for Task 1 at SMM4H 2023. The goal of the task is to automatically distinguish tweets that self-report a COVID-19 diagnosis (for example, a positive test, clinical diagnosis, or hospitalization) from those that do not. We investigate the use of different techniques for preprocessing tweets using four transformer-based models. The ensemble of fine-tuned language models obtained an F1-score of 84.5%, which is 4.1% higher than the average value.*

**Introduction**

Social networks are widely used to exchange news and opinions. The texts in social media are an important source of socially significant information. The automatic analysis of posts provides additional information about personal experiences that people share on social networks.

This work is based on the participation of the team *tmn* in Task 1 – Binary classification of English tweets self-reporting a COVID-19 diagnosis at the Social Media Mining for Health (SMM4H) 2023 workshop [1]. The task aims at automatically classifying tweets that self-report a COVID-19 diagnosis and those that do not. Since the COVID-19 pandemic has had a huge impact on various areas of human life, its consequences are being actively discussed on social media [2]. The task proposed by the organizers of SMM4H is relevant for natural language processing applications that use social media data for health informatics. Since transformer-based models achieve state-of-the-art results in many text classification tasks, we focus on the use of BERT (Bidirectional Encoder Representations from Transformers) [3] and its modifications. We also experiment with several common techniques for preprocessing tweets [4-5].

**Dataset**

The dataset contains 18,000 tweets (7,600 – training subset, 400 – validation, 10,000 – test). 1,391 tweets in the training and validation subsets are labeled as self-reporting, accounting for 17.39% of the dataset. Text length in the dataset varies from 27 to 969 symbols. The average tweet length is 221.63±78.16. Many tweets contain URLs (22.36% among all tweets in the dataset), emoji (53.38%), and mentions (62.21%). A mention means the presence of another person's username anywhere in the body of the tweet.

**Methods**

During the development phase, we compared several pre-trained language models: a) COVID-Twitter-BERT (*CT-BERT*) [6]; b) COVID-Twitter-BERT-v2; c) RoBERTa-large (*RoBERTa*) [7]; d) Twitter-RoBERTa-base (*RoBERTa$_T$*) [8]. We do not provide a technical description of the models due to the restrictions on the paper volume. The relevant papers are listed in References. Using the listed models, we evaluated the impact of tokenization and removal of URLs, mentions, and emoji. Tokenization is a replacement of a part of a tweet with a sequence \$URL\$, \$MENTION\$, and \$EMOJI\$ respectively. For this purpose, we utilized the Preprocessor package[1]. Each model was fine-tuned for two epochs with a maximum sequence length of 128 tokens. A learning rate was equal to 4e-5 for CT-BERT-v2 and RoBERTa$_T$ and 5e-6 for CT-BERT and RoBERTa.

**Results**

We report the results in terms of F1-score (F1) for the class of tweets that self-report the COVID-19 diagnosis (binary averaging). In Table 1, we demonstrate the average scores across three runs for each model on the validation set. The best result was achieved using RoBERTa with tokenizing URLs.

During the test phase, the participants were allowed to submit two sets of predictions. For the first submission, we used a soft-voting ensemble of five RoBERTa models. For the second submission, we utilized an ensemble of all the considered models using a two-step ensemble scheme. First, we obtained the predictions for each model type using a soft-voting technique. We fine-tuned five models for each model type. Second, we applied a hard-voting scheme for final predictions. If the votes are equal, we used the prediction of RoBERTa. For CT-BERT and CT-BERT-v2, we utilized raw texts as input. For RoBERTa, we previously tokenized tweets. For RoBERTa$_T$, we used tweets with removed URLs. The official results for the test set are presented in Table 2.

**Table 1.** The results for the development phase. The scores that exceed the baseline are shown in bold. The highest result for each model is highlighted. The best result is marked with an asterisk (*). T – tokenization, R – removal.

---

[1] https://github.com/s/preprocessor

| Model | Raw (baseline) | URLs | | Mentions | | Emoji | | Combinations of successful techniques |
|---|---|---|---|---|---|---|---|---|
| | | T | R | T | R | T | R | |
| CT-BERT | 84.6 | 84.4 | 81.8 | 82.7 | 83.2 | 81.4 | 82.5 | - |
| CT-BERT-v2 | 86.51 | 82.6 | 81 | 85.4 | 82.8 | 84 | 84.1 | - |
| RoBERTa | 84.7 | **86.8*** | 85.8 | **85.2** | 82.2 | **85.6** | 83.3 | **85.4** - URLs (T), ment. (T), emoji (T) <br> **84.9** - URLs (R), ment. (T), emoji (T) |
| RoBERTa$_T$ | 84.8 | 83.2 | **86.4** | 84.2 | 84.2 | 83.6 | 83.4 | - |

**Table 2.** Official results, %.

| Model | Precision | Recall | F1 |
|---|---|---|---|
| Submission 1 (RoBERTa) | 79.8 | 82.2 | 81 |
| Submission 2 (four models) | 83.3 | 85.8 | 84.5 |
| Average scores provided by the organizers | 82.4 | 79.2 | 80.4 |

**Discussion and Conclusions**

The results presented above show that the ensemble of different language models outperforms the ensemble of single-type models. We also show that some preprocessing techniques, such as tokenizing URLs, mentions, and emoji, can improve the results. However, the improvement is mainly observed for the general-purpose model (RoBERTa) but not for domain-specific models. False positive error examples include citations (*"I ended up in hospital with Covid. This gave me opportunities to tell several nurses that there were people praying for me all around the world, and explain something of what that says about God's family, the church."https://t.co/9o2la81F96*) and the texts that report that another person was diagnosed (*Well Covid's hit abit to close to home for my liking. Mum's tested positive despite having the Vaccine 3 weeks ago*). False negative errors are often related with the tweets that contain too many details specific for the subject area (*@RaeDiamond Bad ass I'm a rare blood type O+, so I feel ya, I had antibodies so I sold some of my plasma in hospital to help a few people, but I was screened and told my antibodies for covid protein were still high, so they said I wasn't a candidate for the vaccine.. that's why I asked*). To avoid these errors, different pretraining approaches and other preprocessing techniques can be explored in further research.

**References**
1. Klein AZ, Banda JM, Guo Y, Flores Amaro JI, Rodriguez-Esteban R, Sarker A, Schmidt AL, Xu D, Gonzalez-Hernandez G. Overview of the eighth Social Media Mining for Health Applications (#SMM4H) Shared Tasks at the AMIA 2023 Annual Symposium. In: Proceedings of the Eighth Social Media Mining for Health Applications (#SMM4H) Workshop and Shared Task; 2023.
2. Glazkova A, Glazkov M, Trifonov T. g2tmn at Constraint@AAAI2021: Exploiting CT-BERT and Ensembling Learning for COVID-19 Fake News Detection. In: Combating Online Hostile Posts in Regional Languages during Emergency Situation. CONSTRAINT 2021, 2021.
3. Devlin J, Chang MV, Lee K, Toutanova K. BERT: Pre-training of Deep Bidirectional Transformers for Language Understanding. In: Proceedings of the 2019 Conference of the North American Chapter of the Association for Computational Linguistics: Human Language Technologies, Vol. 1, 2019.
4. Naseem U, Razzak I, Eklund PW. A survey of pre-processing techniques to improve short-text quality: a case study on hate speech detection on Twitter. Multimedia Tools and Applications. 2021; 80:35239-35266.
5. Symeonidis S, Effrosynidis D, Arampatzis A. A comparative evaluation of pre-processing techniques and their interactions for twitter sentiment analysis. Expert Systems with Applications. 2018; 110:298-310.
6. Müller M, Salathé M, Kummervold PE. COVID-Twitter-BERT: A natural language processing model to analyse COVID-19 content on Twitter. Frontiers in Artificial Intelligence. 2023; 6:1023281.
7. Liu Y, Ott M, Goyal N, Du J, Joshi M, Chen D, Levy O, Lewis M, Zettlemoyer L, Stoyanov V. RoBERTa: A robustly optimized BERT pretraining approach. arXiv preprint arXiv:1907.11692, 2019.
8. Barbieri F, Camacho-Collados J, Anke LE, Neves L. TweetEval: Unified Benchmark and Comparative Evaluation for Tweet Classification. In: Findings of the Association for Computational Linguistics: EMNLP 2020, 2020.